\documentclass[10pt,twocolumn,letterpaper]{article}

\usepackage[pagenumbers]{cvpr} 
\usepackage{tabularx}
\usepackage{booktabs}
\usepackage{xcolor}
\usepackage{pifont}
\usepackage{threeparttable} 
\usepackage{multirow}
\usepackage{adjustbox}
\usepackage{wrapfig}  

\definecolor{cvprblue}{rgb}{0.21,0.49,0.74}
\usepackage[pagebackref,breaklinks,colorlinks,allcolors=cvprblue]{hyperref}

\def\paperID{*****} 
\def\confName{CVPR}
\def\confYear{2026}

\title{RealDrag: The First Dragging Benchmark with Real Target Image}

\author{
Ahmad Zafarani$^1$, Zahra Dehghanian$^1$, Mohammadreza Davoodi$^2$, \\
Mohsen Shadroo$^3$, MohammadAmin Fazli$^1$, Hamid R. Rabiee$^1$ \\[2mm]
$^1$Sharif University of Technology \\
$^2$Iran University of Science and Technology \\
$^3$Comprehensive University of Islamic Revolution \\[2mm]
{\tt\small \{ahmad.zafarani, zahra.dehghanian97, fazli, rabiee\}@sharif.edu} \\
{\tt\small davoodi\_m@comp.iust.ac.ir, mshadroo@cuir.ac.ir}
}
\begin{document}
\maketitle
\begin{abstract}

The evaluation of drag based image editing models is unreliable due to a lack of standardized benchmarks and metrics. This ambiguity stems from inconsistent evaluation protocols and, critically, the absence of datasets containing ground truth target images, making objective comparisons between competing methods difficult. To address this, we introduce \textbf{RealDrag}, the first comprehensive benchmark for point based image editing that includes paired ground truth target images. Our dataset contains over 400 human annotated samples from diverse video sources, providing source/target images, handle/target points, editable region masks, and descriptive captions for both the image and the editing action.

We also propose four novel, task specific metrics: Semantical Distance (SeD), Outer Mask Preserving Score (OMPS), Inner Patch Preserving Score (IPPS), and Directional Similarity (DiS). These metrics are designed to quantify pixel level matching fidelity, check preservation of non edited (out of mask) regions, and measure semantic alignment with the desired task. Using this benchmark, we conduct the first large scale systematic analysis of the field, evaluating 17 SOTA models. Our results reveal clear trade offs among current approaches and establish a robust, reproducible baseline to guide future research. Our dataset and evaluation toolkit will be made publicly available.
\end{abstract}    
\section{Introduction}
\label{sec:intro}

Recent breakthroughs in generative models have completely changed how we edit images \cite{zhang2023adding}. One very practical and easy-to-use method is 'drag-based' editing \cite{pan2023drag,mou2023dragondiffusion}. In this task, a user gives simple instructions by picking handle points and target points. The model job is to update the image so that the handle points move to the target positions while preserving realism \cite{mou2023dragondiffusion}. 
In addition, many methods allow the user to provide masks to indicate which regions should be edited and which regions should remain unchanged, giving finer control over the editing process \cite{mou2023dragondiffusion,nie2023blessing}.

 In this rapidly advancing field, new methods frequently claim state of the art (SOTA) results, often highlighting the limitations of previous work. This makes it challenging to follow true progress, as there is no strong, fair benchmark to provide an unbiased comparison of these competing claims.
 This limitation is two fold problem, first, absence of a balanced, well curated dataset, that has both source and target images so the comparison can be quantified fairly. Second, absence of a task specific metric that can reliably measure the alignment between the manipulated handle points and their target positions while also accounting for realism and semantic consistency. Without these requirements, evaluating and comparing drag based methods remains inconsistent, which hinders reproducibility and systematic benchmarking.

To address these limitations, we introduce RealDrag, the first dedicated dataset for the drag task, consisting of more than 400 carefully curated samples. Each sample includes both handle points and their corresponding target points, enabling a consistent evaluation across methods. Unlike prior works that rely on manually selected or incomplete annotations, our dataset provides a standardized resource for training and benchmarking drag based frameworks fairly. To address second challenge, we propose novel task specific metrics designed to measure both geometric alignment and visual fidelity, offering a reliable basis for quantitative comparison. Finally, using this dataset and evaluation protocol, 
we perform a comprehensive benchmark of existing approaches and provide the first systematic analysis of their strengths and weaknesses for drag-based editing tasks.

\newcommand{\cmark}{\textcolor{green!60!black}{\ding{51}}}
\newcommand{\xmark}{\textcolor{red}{\ding{55}}}

\section{Related Works}
\label{sec:relatedworks}
To build a systematic benchmark for the drag task, we first need to review the field's progress. This section surveys what has happened so far, creating a clear picture of the current landscape. We will start by tracing the evolution of drag-based models, from the foundational methods to the latest SOTA, focusing on their key ideas and novelties. After covering the models, we will then examine the datasets and evaluation metrics that have been proposed for this task.

\subsection{Models}
\label{sec:models}
\begin{figure}
  \centering
  \includegraphics[width=\linewidth]{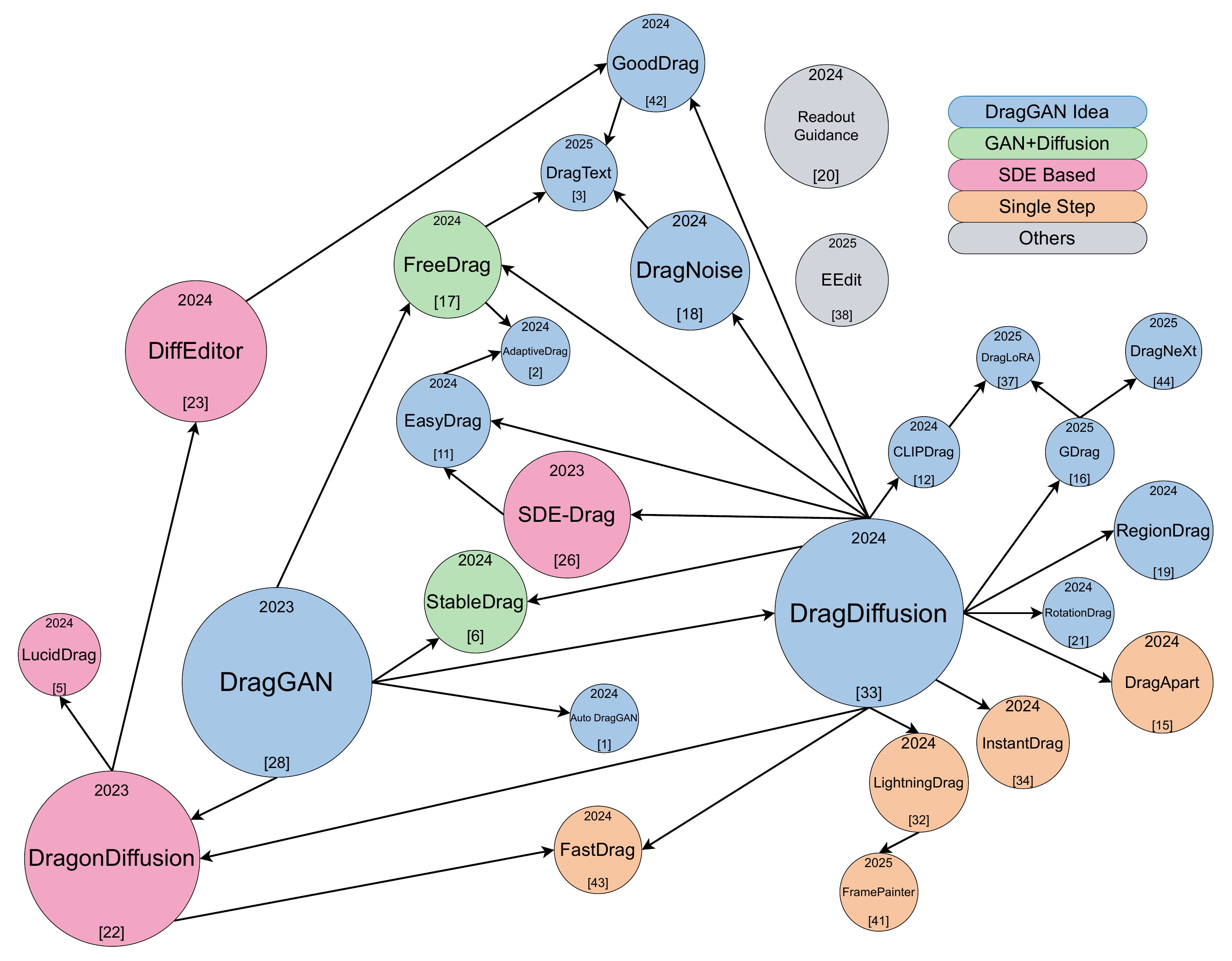}
  \caption{A visualization of the drag-based editing landscape discussed in Section \ref{sec:models}. The node size for each model is proportional to its citation count. Arrows indicate the conceptual hierarchy and progression of ideas. Colors group models by their core methodology. Zoom in to check details.}
  \label{fig:relatedwork}
\end{figure}

Figure \ref{fig:relatedwork} provides a high level map of the drag-based editing field, charting the core ideas and their relative impact that we will discuss here. This progression began with early point-based manipulation, which was framed as a feature space alignment problem. The seminal GAN approach DragGAN\cite{pan2023drag}, introduced this by optimizing latent codes to match handle and target patches, using motion supervision and feature tracking to ensure stable correspondence. Diffusion-based methods soon ported this core concept to denoising UNets, branching into several main strategies. Some methods apply DragGAN style supervision directly to noisy latents, often incorporating attention-based mechanisms to preserve object identity \cite{shi2024dragdiffusion}. Other concurrent branches explored alternatives, such as using learned guidance heads to steer the UNet's inference \cite{luo2024readout} or formulating the drag objective as a feature correspondence energy that is minimized at each denoising step \cite{mou2023dragondiffusion, mou2024diffeditor}.

Subsequent work focused on refining the optimization, exploring what to optimize and when. Instead of continuous latent adjustments, some methods update the bottleneck noise at a specific timestep \cite{liu2024drag} or perturb latents directly and use an SDE to return to the data manifold \cite{nie2023blessing}. A major parallel focus has been on improving stability and mitigating tracker drift. This was addressed with adaptive tracking schemes \cite{ling2024freedrag}, learned tracking modules \cite{cui2024stabledrag}, and more stable supervision strategies that anchor the edit to its start state \cite{zhang2024gooddrag, hou2024easydrag}. This trend also includes specialized solutions for complex motions like rotation \cite{luo2024rotationdrag} and even one step latent prediction in the GAN domain \cite{cai2024auto}.

The push for efficiency and scalability also inspired new strategies. Some methods aim for single shot edits by estimating global latent warps \cite{zhao2024fastdrag}. A significant trend was the shift from point-based to region-level objectives, using segmentation \cite{chen2024adaptivedrag} or handle/target regions \cite{lu2024regiondrag} to define the edit. This also led to entirely alternative formulations, such as learning an optical flow prior to guide the diffusion process \cite{shin2024instantdrag}. To improve generalization, other work focused on developing general plugins \cite{li2024dragapart}. More recently, efforts have aimed to incorporate higher level understanding, like language-based intent \cite{cui2024localize}, or to achieve pure acceleration through token caching and scheduler skipping to cache unedited tokens \cite{yan2025eedit}. To incorporate language, text aligned methods either jointly optimize text embeddings during the drag \cite{choi2025dragtext} or use global CLIP gradients to provide directional guidance \cite{jiang2024clipdrag}. The concept has also been adapted for video, where the edit is treated as a next frame prediction, using video diffusion models to maintain temporal consistency \cite{zhang2025framepainter}.

Finally, the most recent systems have moved towards integrating these ideas into unified controllers. This includes using dual objective LoRA adapters to handle both motion and identity preservation simultaneously \cite{xia2025draglora}. More advanced, task aware controllers can even infer the user's intent (like rotation vs. relocation) and apply a specialized, self adaptive supervision \cite{lin2025gdrag}. Another emerging direction is to move away from point-based optimization, instead using region first controllers that distribute optimization over time for more stable results \cite{zhou2025dragnext}. Table \ref{tab:compare_drag} provides a detailed comparison of the models discussed, including their input types, training status, and parameter counts.

\begin{table}[t]
\centering
\caption{Summary of benchmarked drag editing models and their key characteristics.}
\label{tab:compare_drag}
\small 
\resizebox{\linewidth}{!}{%
\begin{tabular}{@{}p{2.5cm} c c c c c@{}}
\toprule
\textbf{Model} & \multicolumn{2}{c}{\textbf{Interaction}} & \textbf{Code} & \textbf{Training Status} & \textbf{\#Param} \\
\cmidrule(lr){2-3}
 & \textbf{Mask} & \textbf{Text} & & & \\
\midrule
DragGAN\cite{pan2023drag} & \cmark & \xmark & \cmark & Optimization & 0.09B \\
DragDiffusion\cite{shi2024dragdiffusion} & \cmark & \cmark & \cmark & Optimization & 2.1B \\
DragonDiffusion\cite{mou2023dragondiffusion} & \cmark & \cmark & \cmark & Training-free & 1.7B\\
DiffEditor\cite{mou2024diffeditor} & \cmark & \cmark & \cmark & Training-free & 1.7B \\
ReadoutGuidance\cite{luo2024readout} & \xmark & \cmark & \cmark & Optimization & 9.8B \\
DragNoise\cite{liu2024drag} & \xmark & \cmark & \cmark & Optimization & 2.1B \\
SDE-Drag\cite{nie2023blessing} & \cmark & \cmark & \cmark & Training-free & 2.1B\\
FreeDrag\cite{ling2024freedrag} & \cmark & \cmark & \cmark & Optimization & 2.1B \\
GoodDrag\cite{zhang2024gooddrag} & \cmark & \cmark & \cmark & Optimization & 2.1B \\
EasyDrag\cite{hou2024easydrag} & \cmark & \cmark & \cmark & Optimization & 1.4B \\
RotationDrag\cite{luo2024rotationdrag} & \cmark & \cmark & \cmark & Optimization & 2.1B \\
StableDrag\cite{cui2024stabledrag} & \cmark & \xmark & \xmark & Optimization & -- \\
Auto DragGAN\cite{cai2024auto} & \cmark & \xmark & \xmark & Encoder-based & -- \\
FastDrag\cite{zhao2024fastdrag} & \cmark & \cmark & \cmark & Training-free & 2.1B\\
AdaptiveDrag\cite{chen2024adaptivedrag} & \xmark & \cmark & \cmark & Optimization & 2.1B \\
InstantDrag\cite{shin2024instantdrag} & \xmark & \xmark & \cmark & Encoder-based & 0.9B \\
DragText\cite{choi2025dragtext} & \cmark & \cmark & \cmark & Optimization & 2.1B \\
RegionDrag\cite{lu2024regiondrag} & \xmark & \xmark & \cmark & Training-free & 1.7B \\
DragAPart\cite{li2024dragapart}& \xmark & \xmark & \cmark & Encoder-based & 2.0B \\
LucidDrag\cite{cui2024localize} & \cmark & \cmark & \xmark & Optimization & -- \\
LightningDrag\cite{shi2024lightningdrag} & \cmark & \xmark & \cmark & Encoder-based & 2.7B \\
GDrag\cite{lin2025gdrag} & \cmark & \cmark & \xmark & Optimization & -- \\
EEdit\cite{yan2025eedit} & \cmark & \cmark & \cmark & Training-free & 33B \\
CLIPDrag\cite{jiang2024clipdrag} & \cmark & \cmark & \cmark & Optimization & 2.1B \\
FramePainter\cite{zhang2025framepainter} & \xmark & \xmark & \cmark & Encoder-based & 2.2B \\
DragLoRA\cite{xia2025draglora} & \cmark & \cmark & \cmark & Optimization & 2.2B \\
DragNeXt\cite{zhou2025dragnext} & \cmark & \xmark & \xmark & Optimization & -- \\
\bottomrule
\end{tabular}
} 
\end{table}

This progression shows a clear evolution: from GAN latents to diffusion inversion, from simple point-based goals to complex region-, text-, and task-aware objectives, and from basic heuristics to integrated, learned controllers. As we've seen, models now compete along several core design axes: (i) how they estimate correspondence, (ii) how they inject constraints, (iii) their optimization strategy, and (iv) the overall scope of the edit. This complex and rapidly diversifying landscape is precisely why a systematic, fair benchmark is needed to understand the trade-offs and true performance of these varied approaches.

\subsection{Datasets}

Just as the models have rapidly evolved, the datasets used to train and test them have also coevolved. A model's performance is inseparable from the data it's benchmarked on. To understand the current evaluation landscape and its gaps, in this subsection we review the progression of these datasets along three main axes: (i) their drag annotation granularity (from points to regions), (ii) the supervision signals they offer (like prompts and masks), and (iii) their content provenance (real vs. synthetic).

Following that first axis of granularity, the earliest datasets concentrated on point handles. The benchmark released with DragDiffusion \cite{shi2024dragdiffusion} was a key example. It assembled 205 images, providing handle/target pairs and masks. A notable limitation, however, was that its text prompts described the input scene, not the editing action itself. In parallel, the SDE-Drag \cite{nie2023blessing} benchmark introduced a 100 image set with diverse content, including AI generated images, and offered similar point-based annotations. Although both were presented as first benchmarks, they are better understood as two variations of the same point-based approach, differing mainly in their scale and content.

This line of research on annotation granularity was then extended from points to regions. The key example here is RegionDrag \cite{lu2024regiondrag}. Instead of creating a new dataset, this work relabeled the existing SDE-Drag and DragDiffusion benchmarks. It replaced the single handle/target points with more precise polygon or brush regions, creating the new DragBench-SR and DragBench-DR datasets. These new benchmarks offered structured region pairs while importantly preserving the original prompts and masks. By also providing a more consistent mask format, this effort improved reproducibility. In effect, RegionDrag was a targeted extension that made it possible to evaluate modern, region-aware models.

After these initial collections, the focus shifted to the other axes we mentioned: supervision signals and task variety. GoodDrag’s Drag100 \cite{zhang2024gooddrag}, for example, expanded the types of drag tasks. At the same time, it reduced supervision by removing text prompts, which focused the evaluation purely on geometric changes rather than language understanding. A similar trade off was made for GAN specific workflows. FreeDragBench \cite{ling2024freedrag} scaled up to 600 samples but intentionally omitted both prompts and masks. This made it useful for testing latent space editors but less relevant for the many diffusion methods that rely on this external guidance. The theme of these datasets is clear: they traded linguistic/contextual supervision for a wider range of spatial tasks. The priority became robustness across different types of edits, rather than text image grounding.

Finally, we consider the third axis: content provenance, specifically through massive synthetic datasets. A notable instance is DragAPart's Drag-A-Move \cite{li2024dragapart}, which used a 3D part dataset GAPartNet \cite{geng2023gapartnet}, to synthesize an enormous 40 million sample dataset. Unlike the image centric datasets before it, this corpus provides ground truth motion. However, while this provides perfect ground truth, it also introduces significant challenges. Its purely synthetic nature creates a simulation to real gap, making the output of models trained on it look unnatural. More importantly, its focus is narrow: it is designed only for articulated part motion, making it unsuitable for the more common, free form appearance edits that define most drag-based tasks. The details and characteristics of the drag datasets discussed in this section are summarized in Table \ref{tab:datasets_summary}.
Table \ref{tab:datasets_summary} compares existing drag benchmarks against our proposed RealDrag dataset. It highlights that RealDrag is the only benchmark providing real paired target images together with full annotations including masks, captions, and human action descriptions. Earlier datasets either focus on synthetic data, limited annotations, or narrow motion categories. This comparison establishes RealDrag as the first complete and unified dataset capable of supporting consistent evaluation and training across diverse drag scenarios.

\begin{table*}[t]
\centering
\caption{Feature comparison of drag editing datasets. RealDrag (Ours) is the only dataset to provide a complete set of annotations, notably paired target images.}
\label{tab:datasets_summary}
\begin{threeparttable}
\small
\resizebox{\textwidth}{!}{%
\begin{tabular}{@{}p{4.2cm} c c c c c c c c c c@{}}
\toprule
\multirow{2}{*}{\textbf{Dataset Name}} & \multirow{2}{*}{\textbf{\#Samples} }&
\multicolumn{3}{c}{\textbf{Sample Types}} &
\multirow{2}{*}{\textbf{Target}} & \multirow{2}{*}{\textbf{Image}} &
\multirow{2}{*}{\textbf{Action}} & \multirow{2}{*}{\textbf{Mask}} &
\multirow{2}{*}{\textbf{Human}} & \multirow{2}{*}{\textbf{Drag}} 
\\
\cmidrule(lr){3-5}
 & & \textbf{Real} & \textbf{Generated} & \textbf{Animated} &\textbf{Image} & \textbf{Desc.}& \textbf{Prompt}& & \textbf{Evaluation}& \textbf{Category} \\
\midrule
DragBench (DragDiffusion)\cite{shi2024dragdiffusion} & 205 &
\cmark & \cmark & \xmark & \xmark & \cmark & \xmark & \cmark & \cmark & \xmark \\
DragBench (SDE-Drag)\cite{nie2023blessing} & 100 &
\cmark & \cmark & \xmark & \xmark & \cmark & \xmark & \cmark & \cmark & \xmark \\
Drag100 \cite{zhang2024gooddrag} & 100 &
\cmark & \cmark & \xmark & \xmark & \xmark & \xmark & \cmark & \cmark & \xmark \\
FreeDragBench\cite{ling2024freedrag} & 600 &
\xmark & \cmark & \xmark & \xmark & \xmark & \xmark & \xmark & \xmark & \xmark \\
RegionDrag \cite{lu2024regiondrag} & 305\tnote{*} &
\cmark & \cmark & \xmark & \xmark & \cmark & \xmark & \cmark & \cmark & \xmark \\
Drag-A-Move \cite{li2024dragapart} & 40{,}000{,}000 &
\xmark & \cmark & \xmark & \xmark & \xmark & \xmark & \xmark & \xmark & \xmark \\
\midrule
RealDrag (Ours) & 415 &
\cmark & \cmark & \cmark & \cmark & \cmark & \cmark & \cmark & \cmark & \cmark \\
\bottomrule
\end{tabular}%
} 
\begin{tablenotes}
\footnotesize
\item [*] Combination of previous DragBench versions \cite{shi2024dragdiffusion,nie2023blessing}. 
\end{tablenotes}
\end{threeparttable}
\end{table*}

Across all these resources, this review shows that several critical issues remain. Existing real image datasets are small and suffer from category imbalance. Inconsistencies also exist in prompt usage, where often only a description of the scene exists, but an exact prompt for the drag task does not. While region relabeling helped improve spatial precision and synthetic data provided ground truth for a specific category of drag task, a single, unified dataset for all drag tasks is still absent. There is no benchmark that couples clear constraints, standard formats, and explicit action instructions with a diverse set of paired, real world images. To close these exact gaps, our dataset is designed to provide the first fair, direct comparison of drag based methods within a comprehensive and standardized framework.

\subsection{Metrics}

Given the historical absence of ground truth target images in drag-based editing, evaluation has primarily been a relative process. This constraint naturally shaped the development of metrics, which can be broadly classified into two main families: Task Specific and General Purpose.

\paragraph{Task Specific Metrics}
This first family of metrics assesses performance directly against the only two available inputs: the source image and the drag instruction. Their development was driven by the need to answer two fundamental questions based on these inputs:
\begin{itemize}
    \item \textbf{Edit Precision:} 
    This first task specific category assesses how faithfully the model followed the user's instruction. Lacking a ground truth target image, these metrics must indirectly measure precision by quantifying the correspondence between the dragged content in the final output and the target point provided by the user, essentially measuring how precisely the handle points land on their defined targets. This approach began with the Mean Euclidean Distance in GAN-based face editing, which relied on stable facial landmark detectors to compute the error \cite{pan2023drag, mou2023dragondiffusion, cai2024auto, mou2024diffeditor}. To generalize beyond faces, this protocol evolved into feature space matching. A key development in this area is DIFT tracking \cite{tang2023emergent}, which localizes the moved handle by matching features from intermediate diffusion steps. This method is often refined by constraining the feature search to a small mask to improve stability during large deformations \cite{lu2024regiondrag, xia2025draglora}. Finally, to better capture structural relocation and reduce the noise inherent in single point tracking, patch-based metrics were introduced. A notable example is the Dragging Accuracy Index \cite{zhang2024gooddrag}, which averages the error between patches centered at the source and target locations.
    
    \item \textbf{Content Preservation:} The second task specific category, addresses how much of the original image was properly maintained. These metrics try to quantify how well the output maintains the realism of the edited image, its faithfulness to the source content, and critically, the preservation of unchanged regions. To measure this, Perceptual Fidelity metrics like LPIPS, SSIM, and PSNR \cite{Zhang2018TheUE} are widely used. These quantify low level structural similarity and perceptual differences between the original and edited images, which is particularly important for ensuring non edited regions are not degraded \cite{ling2024freedrag, shin2024instantdrag, choi2025dragtext, lu2024regiondrag, li2024dragapart, lin2025gdrag, xia2025draglora, zhang2025framepainter}. Beyond direct comparison, Distributional Realism is assessed using Frechet Inception Distance (FID) \cite{parmar2021cleanfid}. This metric is commonly used to detect unrealistic artifacts by comparing the feature distribution of the edited outputs against that of the original images \cite{pan2023drag, mou2023dragondiffusion, ling2024freedrag, mou2024diffeditor, zhou2025dragnext}. Finally, high level Semantic Alignment is measured with CLIP-based scores. CLIP Similarity \cite{Radford2021LearningTV} ensures the edited image remains semantically consistent with the source \cite{shin2024instantdrag, cui2024localize, xia2025draglora}, while CLIP-FID \cite{Kynkaanniemi2022TheRO} is applied in text guided settings to measure alignment with both image realism and a text prompt \cite{zhang2025framepainter}.
\end{itemize}

\paragraph{General Purpose Metrics}
This second category evaluates properties of the output image and the model that are independent of the specific edit instruction itself. These metrics typically fall into three clusters:
\begin{itemize}
    \item \textbf{Computational Efficiency:}
    This first general purpose category evaluates the practical cost of using a model, making the trade-offs between accuracy and performance explicit. This typically includes Latency, measured as the end-to-end inference time, which is often split into preparation and optimization stages \cite{mou2023dragondiffusion, nie2023blessing, hou2024easydrag, ling2024freedrag, mou2024diffeditor, zhao2024fastdrag, chen2024adaptivedrag, shin2024instantdrag, cui2024localize, cui2024stabledrag, jiang2024clipdrag, xia2025draglora, zhou2025dragnext}. It also covers Memory, specifically the peak GPU consumption required to run the edit \cite{shin2024instantdrag, cui2024localize}. Finally, other Operational Costs are considered, such as the total number of denoising steps or the parameter counts for any trainable adapters \cite{xia2025draglora}.

    \item \textbf{Overall Aesthetic Quality:} 
    This general purpose category aims to capture the subjective visual appeal of the result, addressing aspects of quality that many automated metrics miss. To do this, researchers rely on perceptual judgments, which are gathered in two primary ways. The first is through Human Studies. Early protocols relied on small scale pairwise A/B tests \cite{nie2023blessing}, but this has since evolved into more comprehensive studies where human raters rank models on multiple criteria, such as visual consistency, edit accuracy, and overall quality \cite{mou2023dragondiffusion, zhang2024gooddrag, shin2024instantdrag, lin2025gdrag, luo2024rotationdrag, zhang2025framepainter}. As a high throughput alternative to these costly human studies, vision language model (VLM)-Based Judges have been introduced. A key example is GScore \cite{zhang2024gooddrag, cui2024localize}, which prompts a large VLM to rate the perceptual quality of an edit.
    

\item \textbf{Robustness and Specialized Protocols}
A final general purpose category includes specialized tests designed to probe model behavior and reliability. One such protocol is Paired Reconstruction, primarily used in GANs \cite{cai2024auto}. This tests if a model can accurately synthesize a known target image $I_2$ from a source $I_1$ given sparse points, using MSE/LPIPS as the reconstruction loss \cite{pan2023drag}. Another protocol is Symmetrical Dragging, which serves as a robustness check \cite{ling2024freedrag}. This test applies an edit, inverts the instruction, and applies it again; the LPIPS distance from the final image back to the original penalizes asymmetric or irreversible edits.


\end{itemize}


This evaluation landscape shows a clear progression from domain specific metrics to a comprehensive suite for open domain editing. A modern, holistic benchmark must be grounded in a core set of metrics that capture the fundamental trade offs of the task.

\section{Methodology}
\label{sec:methodology}
As our review in Section \ref{sec:relatedworks} demonstrated, the current evaluation landscape for drag-based editing is fragmented and lacks a unified standard. Key components are missing, most notably the absence of ground truth target images and standardized, drag task specific metrics.

To address these critical gaps, this section introduces our complete methodology. We first present the RealDrag dataset, a novel, large scale benchmark designed to fill these gaps. We will detail its construction, diverse content, and rich annotations. Second, we introduce our proposed evaluation metrics, which are designed to work with our new dataset to provide a more accurate measure of model performance.

\subsection{Dataset}



While there have been many attempts, achieving precise control over one part of an image while keeping other parts unchanged remains a challenging task \cite{li2024controlnet++}. Tasks such as dragging visual elements to new positions fall into this category, and are a poor fit for traditional text-based editing due to it’s inherently ambiguous for geometric reasoning; for example, a text prompt cannot precisely indicate the final position of a hand.

This has led to the development of drag-based editing, but the datasets for this task have severe limitations. First, they often lack semantic richness. Many resources represent an entire manipulation with just a single displacement vector, a format that is fundamentally unclear. For instance, a simple drag vector on a human face could imply a full spatial translation or a subtle rotation, leading to inconsistent model learning. Second, and most critically, publicly available datasets lack paired, ground truth target images. They generally rely on synthetic or automatically generated annotations. This absence of real-world ground truth makes it impossible to reliably benchmark models or supervise training for high-fidelity results.

\begin{figure*}
  \centering
  \includegraphics[width=1\textwidth]{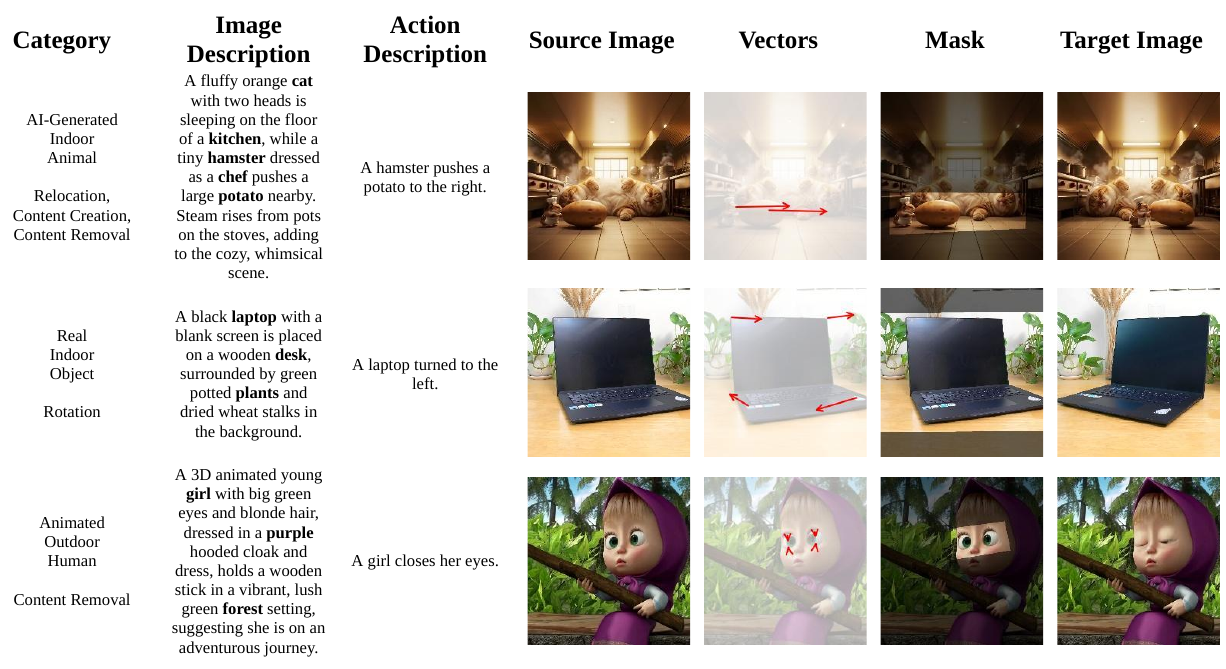}
  \caption{Visual overview of the RealDrag dataset. The dataset provides rich annotations, including Target Image, Image Category, Action description and more.}
  \label{fig:datasetsamples}
\end{figure*}

To address all of these limitations, we construct our new dataset. RealDrag introduces explicit, natural language action prompts to resolve semantic ambiguity. It also provides editable region masks, input image captions, and for the first time, paired ground truth target images. This combination provides the structured supervision necessary for this complex task and enables a fair, consistent evaluation across a diverse range of content. Some samples from our dataset, illustrating these rich annotations, are shown in Figure \ref{fig:datasetsamples}.

\paragraph{Dataset Generation}

To capture realistic drag operations, we built the dataset using a semi automatic pipeline that combines programmatic data collection, automated visual filtering, and final human annotation. First, we gathered a diverse set of source materials. This included videos from established datasets like Charades \cite{sigurdsson2016hollywood}, Animal Kingdom \cite{ng2022animal}, and OpenVid-1M \cite{nan2024openvid}, as well as clips from YouTube \cite{youtube}, Civitai \cite{civitai}, and AI generated content using the Nano Banana model \cite{nanobanana} to ensure variety. To further enhance our collection with high fidelity real images, we also asked a photographer to capture specific real world scenarios for this task.


Next, we programmatically filtered this large collection. To isolate object motion from camera motion, we automatically selected only those clips recorded from a stable camera viewpoint. We identified these clips using a motion stability metric, specifically the Farneback optical flow method \cite{Farnebck2003TwoFrameME}. From each of these stable clips, two representative frames were extracted to serve as the source/input image and the ground truth/target image for the drag operation.


The final stage of our pipeline is annotation, which was conducted in several steps. First, preliminary descriptions were generated with GPT-4.1 \cite{openai2023gpt4} to establish a consistent textual baseline. Following this, these annotations were rigorously refined and validated by two trained human experts. Crucially, our experts wrote the final drag action descriptions and categories entirely manually. This was vital to preserve the precise nuance and intent of the edit, which automated methods often miss. In total, we obtained 415 human annotated samples, forming the largest manually curated dataset for drag-based editing to date.


The resulting dataset is a diverse collection, comprising real, synthetic, and animated samples with various resolutions and a controlled number of modifications per instance. This careful design ensures a balanced distribution of motion types and object categories, which is crucial for maintaining interpretability and reproducibility in down stream model training and evaluation.

\paragraph{Dataset Analysis}

We performed a statistical analysis of our dataset's composition, summarized in Supplementary \ref{sup:dataset analysis}. A key contribution of our work is the introduction of task aware labels. As shown in Figure \ref{fig: datasettasks} (c), our dataset is the first to categorize and labeled all samples into five distinct manipulation types: relocation, rescalation, rotation, content creation, and content removal. Providing these labels allows, for the first time, a way to understand the editor's intent.

Beyond task labels, we also ensured the dataset has broad contextual and semantic diversity, as summarized in Figure \ref{fig: datasettasks} (a). The data is balanced between indoor and outdoor scenes. It also contains a mix of real, animated, and AI generated images, with subjects distributed across humans, animals, and objects. This broad coverage across scene contexts, generation sources, and semantic categories is essential. It ensures that the dataset can support the evaluation of a model's capabilities on specific manipulation tasks and varied visual and structural conditions rather than just a single generic drag score.

\subsection{Metric}

A comprehensive benchmark requires both a high quality dataset and a robust set of metrics. Our approach here is two fold. First, to ground our benchmark in the existing literature, we select a suite of established, general purpose metrics that are widely used to measure aspects like perceptual quality and realism.

Second, we must address a critical gap. Because no previous dataset contained ground truth target images, an entire dimension of evaluation how similar the edited photo is to the real, desired outcome has gone unmeasured. Our dataset is the first to make this measurement possible. Therefore, to capture this new and vital dimension, we propose a set of novel, task specific metrics designed to quantify fidelity, preservation, and semantic alignment against a real ground truth.

\paragraph{Proposed Metrics.}
While the selected conventional metrics offer a general overview, they are limited in describing localized deformation accuracy and semantic fidelity against a real target. We therefore introduce four complementary metrics, enabled by our dataset, to measure this. In the formulas that follow, we use $I_S$ for the source image, $I_G$ for the generated image, and $I_T$ for the ground truth target image. $M$ represents the editable region mask (where pixels to be edited are 1) and $\bar{M}$ is its inverse. $D_A$ is the text-based action description. $P[V, I]$ denotes a patch extracted at point $V$ from an image $I$, and $\odot$ represents element wise multiplication.

\begin{itemize}
    \item \textbf{Semantical Distance (SeD)} measures the perceptual distance between the generated image and the ground truth target, focusing only on the editable region. This captures semantic alignment beyond simple pixel differences. In our work, we implement this using LPIPS, calculated as:
    \begin{equation}
    \text{SeD} = \text{LPIPS}(I_T \odot M, I_G \odot M)
    \end{equation}

    \item \textbf{Directional Similarity (DiS)} quantifies the directional consistency between the intended edit and the actual semantic change. It checks if the semantic change from the source to the generated image aligns with the action description, using a 
    \begin{equation}
    \text{DiS} = |\frac{\text{CS}(E(I_S) - E(I_G), E(D_A))}{\text{CS}(E(I_T) - E(I_G), E(D_A))} - 1|
    \end{equation}
    In this equation $E$ is a joint text image embedding space $E$ such as CLIP \cite{radford2021learning} and CS is a Cosine Similariy between two embeddings.
    
    \item \textbf{Outer Mask Preserving Score (OMPS)} computes the pixel-wise similarity between the non-edited regions of the generated imageand the original source image. This ensures the model performs a localized modification without degrading the background.
    \begin{equation}
    \text{OMPS} = |\frac{\text{RMSE}(I_S \odot \bar{M}, I_G \odot \bar{M})}{\text{RMSE}(I_S \odot \bar{M}, I_T \odot \bar{M})} - 1|
    \end{equation}

    \item \textbf{Inner Patch Preserving Score (IPPS)} evaluates whether the modified region remains structurally consistent with both the source and target patches. It emphasizes fidelity during controlled transformations.
    \begin{equation}
    \text{IPPS} = |\frac{\text{RMSE}(P[V_{\text{head}}, I_G], P[V_{\text{tail}}, I_S])}{\text{RMSE}(P[V_{\text{head}}, I_T], P[V_{\text{tail}}, I_S])} - 1|
    \end{equation}
    where $P[V_{\text{head}}, I]$ and $P[V_{\text{tail}}, I]$ are patches at the handle and target points, respectively.
\end{itemize}

\section{Experiments}
\label{sec:experimnets}

\subsection{Implementation Details}
All experiments were run on a single NVIDIA A100 GPU (80 GB). Each model followed its official inference setup as described in its paper or code. All hyperparameters, including random seed and inference steps, were kept as in the original Gradio or code defaults. When a model included optional inputs such as prompts or masks, we applied them only if recommended by the authors. For example, if a method advised not to use prompts, we omitted them.

Some models were excluded from our benchmark for various reasons:
\textbf{DragGAN}\cite{pan2023drag} lacks a universal model;
\textbf{RegionDrag}\cite{lu2024regiondrag} uses region-based inputs instead of points;
\textbf{DragAPart}\cite{li2024dragapart}'s input is limited to articulated object;
and \textbf{FramePainter}\cite{zhang2025framepainter} does not support drag-based interaction.

\subsection{Quantitative Evaluation}
All quantitative evaluations are conducted on the RealDrag dataset. About 3\% of samples from both tails of the data distribution (6\% in total) are treated as outliers and removed. These samples typically correspond to scenes with unclear motion or inconsistent labels. The mean and standard deviation of each metric are then computed on the remaining data.

\subsubsection{Selected Metrics}

As the first part of our evaluation, we use a suite of established metrics to ensure comparability with prior work. Our criteria for inclusion were metrics that are either (1) widely reported in the literature, or (2) provide crucial insight into model efficiency and visual quality. This foundational suite includes: FID for distribution-level realism; Image Fidelity for reconstruction accuracy; Mean Distance for geometric accuracy of control points; SSIM for low-level structural similarity; and both Inference Time (including LoRA training time for models advised to use it) and GPU memory consumed to quantify operational cost. Table~\ref{tab:previous_metrics} reports the average performance of all models on these metrics.
EEdit \cite{yan2025eedit} performs less extensive edits, which leads to higher SSIM and lower FID scores in a acceptable time, albeit at the cost of significantly higher GPU consumption. DragLoRA \cite{xia2025draglora} achieves better performance on core drag metrics such as mean distance and image fidelity, but requires much longer inference time. In contrast, lightweight models like InstantDrag \cite{shin2024instantdrag} and LightningDrag \cite{shi2024lightningdrag} offer faster execution with noticeable quality degradation.

\begin{table*}[t]
\centering
\caption{Quantitative evaluation across 17 models using our metrics. $\downarrow$ indicates lower is better, and $\uparrow$ indicates higher is better.}
\label{tab:previous_metrics}
\begin{threeparttable}
\small
\setlength{\tabcolsep}{6pt}
\renewcommand{\arraystretch}{1.1}
\resizebox{\linewidth}{!}{%
\begin{tabular}{lcccccc}
\toprule
\textbf{Model} & \textbf{FID} $\downarrow$ & \textbf{SSIM\tnote{*}} $\uparrow$ & \textbf{Image Fidelity\tnote{*}} $\uparrow$ & \textbf{Mean Distance} $\downarrow$ & \textbf{Inference Time (s)} $\downarrow$ & \textbf{Consumed GPU (GB)} $\downarrow$ \\
\midrule
AdaptiveDrag\cite{chen2024adaptivedrag} & 85.66 ± 53.16 & 74.68 ± 11.4 & \textbf{\textcolor{orange}{21.53 ± 12.3}} & 86.27 ± 76.18 & 137.49 ± 39.7 & 12.05 ± 0.12 \\ 
CLIPDrag\cite{jiang2024clipdrag} & \textbf{\textcolor{orange}{54.25 ± 39.62}} & 81.72 ± 9.98 & 11.14 ± 7.7 & 109.83 ± 77.87 & 114.25 ± 51.56 & 16.17 ± 0.01 \\ 
DragDiffusion\cite{shi2024dragdiffusion} & 73.47 ± 45.22 & 78.1 ± 10.33 & 16.4 ± 8.85 & 89.99 ± 76.5 & 127.74 ± 54.22 & 11.79 ± 0.0 \\ 
DragLoRA\cite{xia2025draglora} & 93.43 ± 57.64 & 74.8 ± 11.67 & \textbf{\textcolor{red}{19.3 ± 10.52}} & \textbf{\textcolor{blue}{58.52 ± 48.21}} & 134.33 ± 47.75 & 14.53 ± 0.44 \\ 
DragNoise\cite{liu2024drag} & \textbf{\textcolor{red}{58.83 ± 35.81}} & \textbf{\textcolor{red}{83.88 ± 8.66}} & 8.87 ± 3.72 & 116.55 ± 85.87 & 70.16 ± 32.82 & 13.12 ± 0.01 \\ 
DragonDiffusion\cite{mou2023dragondiffusion} & 114.8 ± 64.66 & 82.55 ± 7.58 & 16.8 ± 7.14 & 76.22 ± 86.56 & 67.11 ± 19.38 & 7.8 ± 0.51 \\ 
DragText\cite{choi2025dragtext} & 81.03 ± 48.82 & 77.54 ± 10.28 & 17.57 ± 9.17 & 84.13 ± 73.61 & 111.95 ± 46.74 & 12.39 ± 0.0 \\
EasyDrag\cite{hou2024easydrag} & 101.54 ± 62.4 & 77.15 ± 10.78 & 17.88 ± 9.46 & 85.41 ± 67.17 & 88.39 ± 54.72 & 26.07 ± 3.55 \\ 
EEdit\cite{yan2025eedit} & \textbf{\textcolor{blue}{19.11 ± 10.59}} & \textbf{\textcolor{blue}{97.3 ± 1.92}} & 1.54 ± 1.06 & 145.82 ± 93.09 & \textbf{\textcolor{red}{38.06 ± 8.31}} & 42.99 ± 2.3 \\ 
FastDrag\cite{zhao2024fastdrag} & 112.44 ± 61.84 & 73.81 ± 11.57 & \textbf{\textcolor{blue}{21.71 ± 10.95}} & 81.17 ± 68.9 & 93.54 ± 52.77 & \textbf{\textcolor{red}{6.7 ± 0.05}} \\ 
FreeDrag\cite{ling2024freedrag} & 73.44 ± 47.14 & 83.36 ± 8.53 & 11.0 ± 5.51 & 120.35 ± 84.46 & 90.31 ± 32.35 & 13.33 ± 0.2 \\ 
GoodDrag\cite{zhang2024gooddrag} & 82.44 ± 49.13 & 76.84 ± 10.46 & 17.61 ± 9.04 & \textbf{\textcolor{red}{74.73 ± 67.71}} & 127.74 ± 44.13 & 13.56 ± 0.03 \\ 
InstantDrag\cite{shin2024instantdrag} & 100.44 ± 67.45 & 72.0 ± 11.53 & 17.05 ± 8.87 & 95.01 ± 85.49 & \textbf{\textcolor{orange}{3.9 ± 2.23}} & \textbf{\textcolor{blue}{2.78 ± 0.0}} \\ 
LightningDrag\cite{shi2024lightningdrag} & 95.2 ± 55.04 & 77.23 ± 10.02 & 18.66 ± 9.57 & \textbf{\textcolor{orange}{61.29 ± 69.11}} & \textbf{\textcolor{blue}{0.4 ± 0.49}} & \textbf{\textcolor{orange}{6.28 ± 0.01}} \\ 
ReadoutGuidance\cite{luo2024readout} & 131.37 ± 66.09 & 78.97 ± 10.05 & 17.69 ± 6.77 & 161.57 ± 101.2 & 135.34 ± 18.29 & 18.09 ± 0.0 \\ 
RotationDrag\cite{luo2024rotationdrag} & 134.12 ± 76.04 & 79.87 ± 9.22 & 18.35 ± 8.32 & 155.72 ± 108.8 & 786.97 ± 476.3 & 11.99 ± 0.12 \\ 
SDE-Drag\cite{nie2023blessing} & 96.46 ± 71.68 & \textbf{\textcolor{orange}{86.34 ± 7.43}} & 9.75 ± 5.71 & 137.91 ± 102.3 & 653.74 ± 500.01 & 7.88 ± 0.53 \\ 

\bottomrule
\end{tabular}
}
\begin{tablenotes}
\footnotesize
\item [*] These metrics are ×$10^2$
\end{tablenotes}
\end{threeparttable}
\end{table*}

In general, these metrics capture overall image realism and efficiency. However, as shown in Figure \ref{fig: qualitative} and other qualitative comparisons in the appendix, the models that appear best to human perception are not necessarily those favored by these metrics.
Their measurement scope remains limited in capturing fine spatial precision and semantic consistency. While they can indicate general performance trends, they fail to reflect localized deformations or task-specific alignment.
This limitation mainly arises from the lack of ground-truth target images, highlighting the need for metrics that explicitly involve real reference outputs during evaluation.

\subsubsection{Proposed Metrics}
Table ~\ref{tab:our_metrics} summarizes the performance of evaluated models under our proposed metrics. The proposed metrics complement each other; for instance, AdaptiveDrag\cite{chen2024adaptivedrag} shows strong directional alignment with user intent (good DiS) but struggles to preserve masked regions (poor OMPS). Similarly, DragDiffusion\cite{mou2023dragondiffusion} often makes target points similar to handle points (high IPPS) but fails to maintain semantic consistency and spatial accuracy (poor SeD and DiS). Ultimately, performing minimal edits seems to be the safest strategy; EEdit\cite{yan2025eedit} ranks among the top models across three metrics not because of exceptional capabilities, but because competing methods often fail at the drag task, producing unrealistic images that deviate even more from the target image than the source image.

\begin{table}[t]
\centering
\caption{Quantitative evaluation across 17 models using our metrics. $\downarrow$ indicates lower is better.}
\label{tab:our_metrics}
\begin{threeparttable}
\small
\resizebox{\linewidth}{!}{%
\begin{tabular}{@{}p{2.5cm} c c c c@{}}
\toprule
\textbf{Model} & \textbf{DiS\tnote{*}} $\downarrow$ & \textbf{IPPS\tnote{*}} $\downarrow$ & \textbf{OMPS\tnote{*}} $\downarrow$ & \textbf{SeD\tnote{*}} $\downarrow$ \\
\midrule
        AdaptiveDrag\cite{chen2024adaptivedrag} & \textbf{\textcolor{blue}{133.86 ± 121.5}} & 10.86 ± 13.31 & 31.15 ± 33.19 & 16.29 ± 11.05 \\
        CLIPDrag\cite{jiang2024clipdrag} & 143.29 ± 136.95 & 12.22 ± 16.74 & 19.11 ± 16.81 & 15.76 ± 11.54 \\
        DragDiffusion\cite{shi2024dragdiffusion} & 141.95 ± 136.39 & 11.15 ± 14.34 & 21.35 ± 20.81 & 15.89 ± 11.16 \\
        DragLoRA\cite{xia2025draglora} & 145.78 ± 147.45 & 12.06 ± 15.14 & 21.46 ± 21.98 & 15.94 ± 11.28 \\
        DragNoise\cite{liu2024drag} & 146.8 ± 135.24 & 11.84 ± 15.86 & \textbf{\textcolor{blue}{16.78 ± 15.85}} & \textbf{\textcolor{orange}{15.35 ± 11.08}} \\
        DragonDiffusion\cite{mou2023dragondiffusion} & 141.04 ± 148.81 & \textbf{\textcolor{blue}{9.87 ± 9.69}} & 25.73 ± 29.13 & 17.17 ± 11.35 \\
        DragText\cite{choi2025dragtext} & 147.94 ± 140.99 & 11.19 ± 14.15 & 20.75 ± 20.09 & 15.77 ± 11.22 \\
        EasyDrag\cite{hou2024easydrag} & 141.86 ± 118.11 & 11.42 ± 14.46 & 27.56 ± 27.99 & 16.63 ± 11.59 \\
        EEdit\cite{yan2025eedit} & \textbf{\textcolor{red}{137.31 ± 106.22}} & \textbf{\textcolor{orange}{10.23 ± 12.68}} & 43.28 ± 16.71 & \textbf{\textcolor{blue}{14.93 ± 11.07}} \\
        FastDrag\cite{zhao2024fastdrag} & 140.23 ± 127.05 & \textbf{\textcolor{red}{10.61 ± 12.49}} & 21.33 ± 21.74 & 17.40 ± 11.87 \\
        FreeDrag\cite{ling2024freedrag} & 154.56 ± 149.68 & 11.84 ± 15.86 & 20.4 ± 21.12 & 15.90 ± 11.44 \\
        GoodDrag\cite{zhang2024gooddrag} & 155.26 ± 160.34 & 11.38 ± 13.53 & 23.76 ± 23.29 & 16.26 ± 11.30 \\
        InstantDrag\cite{shin2024instantdrag} & 153.58 ± 164.2 & 11.43 ± 13.52 & \textbf{\textcolor{red}{18.76 ± 19.62}} & 16.15 ± 11.30 \\
        LightningDrag\cite{shi2024lightningdrag} & 143.31 ± 149.56 & 11.69 ± 13.74 & 27.09 ± 25.23 & \textbf{\textcolor{red}{15.68 ± 11.53}} \\
        ReadoutGuidance\cite{luo2024readout} & \textbf{\textcolor{orange}{136.46 ± 120.31}} & 12.95 ± 17.19 & 40.54 ± 36.74 & 18.32 ± 12.21 \\
        RotationDrag\cite{luo2024rotationdrag} & 160.71 ± 149.82 & 11.91 ± 16.04 & 23.3 ± 22.92 & 18.24 ± 12.29 \\
        SDE-Drag\cite{nie2023blessing} & 144.9 ± 139.11 & 11.56 ± 13.86 & \textbf{\textcolor{orange}{17.89 ± 14.78}} & 16.56 ± 11.80 \\
    \bottomrule
\end{tabular}
} 
\begin{tablenotes}
\footnotesize
\item [*] These metrics are ×$10^2$
\end{tablenotes}
\end{threeparttable}
\end{table}

Compared to standard measures, these metrics better expose differences in spatial precision and semantic consistency, and it shows the effectiveness of our proposed Metrics.

\subsection{Qualitative Evaluation}

\label{sec: qualitative}
We provide a huge qualitative comparisons in Supplementary \ref{sup: qualitative}, where each row shows the source and target images followed by the outputs of different models. The target image serves as ground truth to visualize how closely each method aligns with the desired edit.

\section{Conclusion}
In this work, we present \textbf{RealDrag}, the first comprehensive benchmark for evaluating drag-based image editing with paired real target images. Through the construction of a balanced and diverse dataset with detailed annotations, we addressed long standing limitations in existing benchmarks that relied on incomplete or synthetic supervision. Our dataset provides not only paired ground truth targets but also explicit action descriptions and editable region masks, enabling a fair and standardized evaluation across models.

To complement this dataset, we introduced four novel metrics: Semantical Distance, Directional Similarity, Outer Mask Preserving Score, and Inner Patch Preserving Score. to capture geometric accuracy, content preservation, and semantic consistency. Using these metrics, we performed the first large scale systematic analysis across all available SOTA drag models, establishing clear baselines and insights into their trade offs.

\label{sec:conclusion}

{
    \small
    \bibliographystyle{ieeenat_fullname}
    \bibliography{main}
}

\clearpage
\setcounter{page}{1}
\maketitlesupplementary
\appendix

\section{Dataset Analysis}
\label{sup:dataset analysis}

\begin{figure*}
  \centering
  \includegraphics[width=0.75\textwidth]{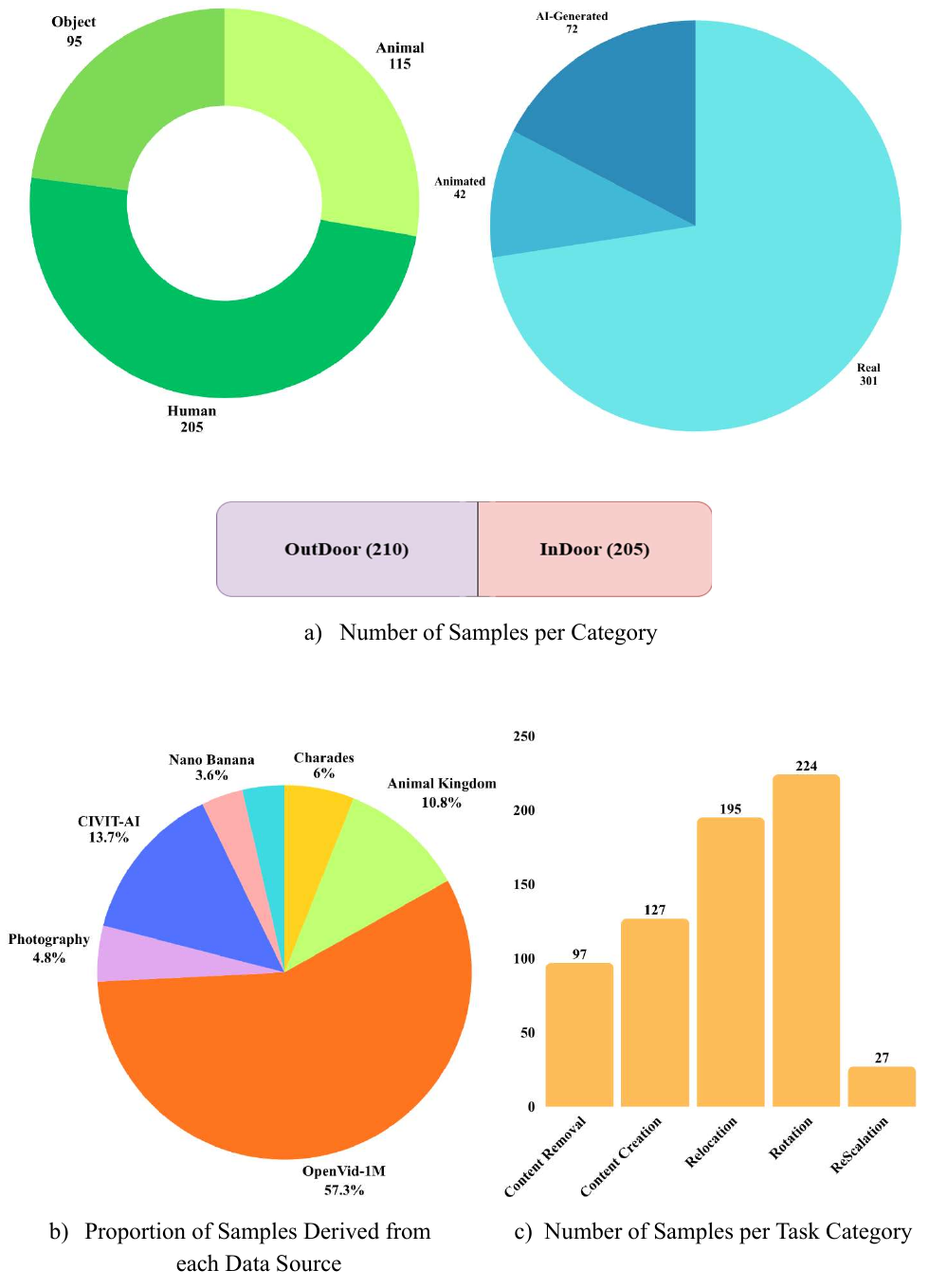}
  \caption{Dataset Analysis}
  \label{fig: datasettasks}
\end{figure*}

This section provides a detailed overview of our dataset through a set of descriptive figures. Figure \ref{fig: datasettasks} (a) presents the distribution of samples across the dataset’s high-level categories (indoor/outdoor, real/animated/AI-generated, and animal/human/object). Figure \ref{fig: datasettasks} (b) reports the proportion of samples originating from each data source used to construct the dataset. Figure \ref{fig: datasettasks} (c) illustrates the distribution of samples across drag-task categories. Note that task categories are not mutually exclusive—a single sample may belong to multiple task types.

\section{Qualitative Results}
\label{sup: qualitative}

\begin{figure*}
  \centering
  \includegraphics[width=1\textwidth]{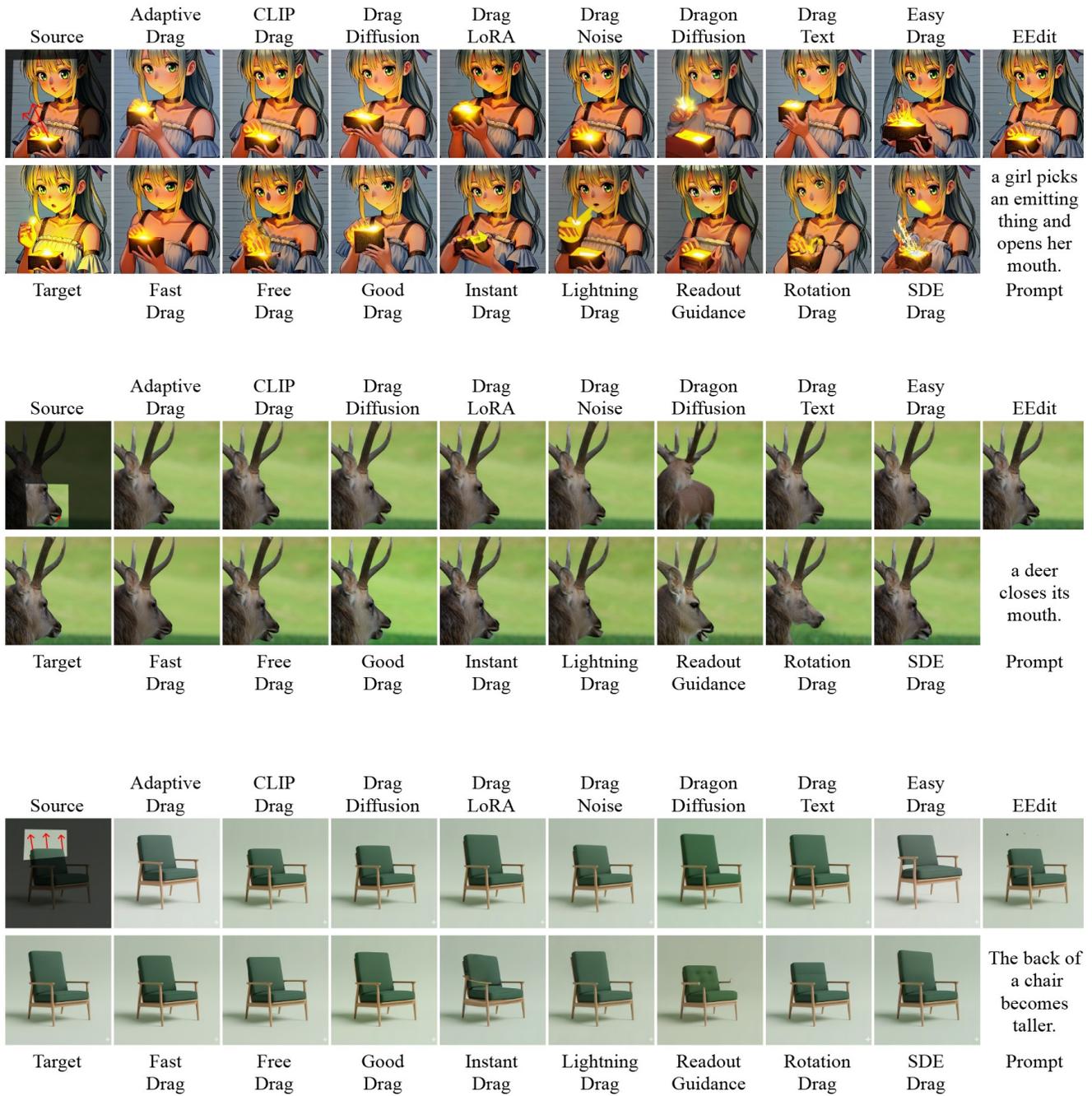}
  \caption{Qualitative Results}
  \label{fig: qualitative}
\end{figure*}

\begin{figure*}
  \centering
  \includegraphics[width=1\textwidth]{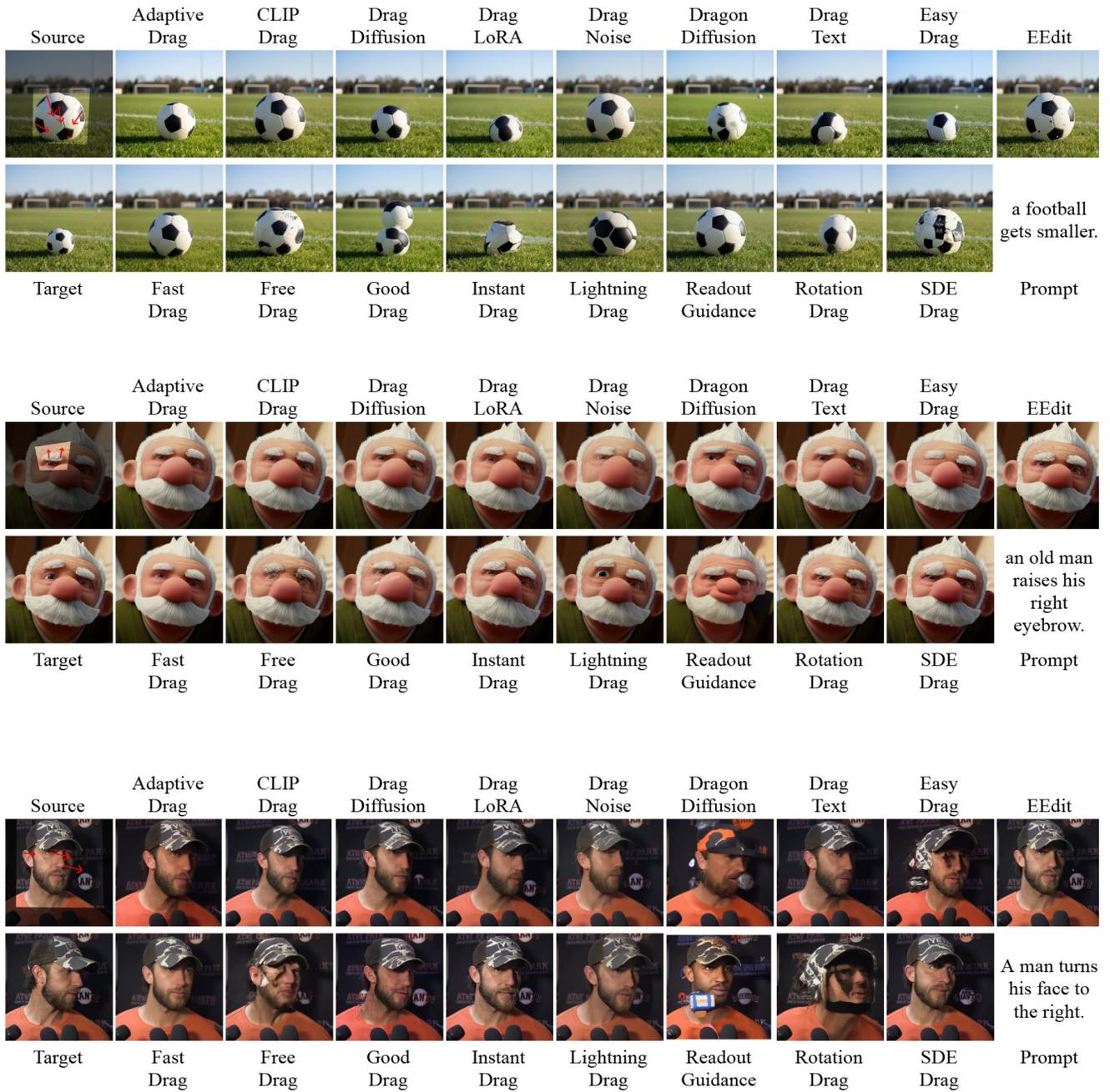}
  \caption{More Qualitative Results}
  \label{fig: qualitative_2}
\end{figure*}

In this section, we present extensive qualitative results to facilitate a clearer and more intuitive comparison among the evaluated models. The visual examples in Figures \ref{fig: qualitative} and \ref{fig: qualitative_2} highlight the characteristic behaviors, strengths, and failure modes of each method, complementing the quantitative analyses reported in previous sections. Note that the prompt shown for each image set corresponds to the action prompt, which is only supported by CLIPDrag \cite{jiang2024clipdrag}.


\section{Proposed Metrics: Motivation and Analysis}
\label{sup: metrics}

This section provides an in-depth examination of our proposed evaluation metrics, outlining the motivation behind their formulation and the specific aspects of the drag task they are intended to capture. We describe how the metrics collectively cover complementary dimensions of model behavior—ranging from semantic alignment to spatial precision and preservation of unedited regions—and why each component is necessary for a comprehensive assessment.
The following subsections elaborate on these ideas from three perspectives:
(C.1) a conceptual discussion of the design choices underlying each metric,
(C.2) a task-wise evaluation of model performance across the five drag task types, and
(C.3) an ablation analysis of the IPPS metric that examines the sensitivity of its behavior to different patch sizes.
Together, these analyses offer a clearer understanding of how the metrics characterize model performance and justify their use in the main evaluation framework.

\subsection{Discussion}
In the metrics presented in this paper, the emphasis was on leveraging the capabilities afforded by the new \textbf{RealDrag} dataset. For instance, in the SeD metric, we sought to evaluate the similarity of the editable region (within the mask) to the ground truth image. Pixel-wise similarity proved overly stringent for this purpose; therefore, we eschewed metrics such as MSE or RMSE in favor of a perceptual metric like LPIPS.

The DiS metric draws upon established principles from the image-to-image translation task. Our objective was to assess the alignment between the modifications in the generated image relative to the input image and the action prompt. To achieve this, we employed the CLIP model to compare text and image embeddings in a shared latent space. In this metric, as well as the subsequent two metrics, the availability of a ground truth image enabled us to benchmark the computed metric against an idealized outcome. Consequently, these metrics feature a denominator analogous to the numerator, but substituting the target image for the generated one. This form of division (normalization) offers a key advantage: it facilitates cross-sample comparisons. However, the optimal value for such a metric is 1, with deviations above or below being undesirable; thus, we subtracted the fraction from 1 and applied the absolute value to frame values closer to zero as more favorable.

In the OMPS metric, the rigorous RMSE is employed, as models that incorporate the mask as input should induce no alterations whatsoever in the unmasked region.

The IPPS metric aims to compare a patch surrounding the endpoint of the drag vector with its starting point. In the drag editing task, these two regions are expected to exhibit similarity (hence our use of RMSE). Moreover, owing to the normalization described, we can assert that this metric is fair. In subsection C.3, we performed an ablation study on the size of this patch.

\subsection{Per Task Evaluations}
\begin{figure*}
  \centering
  \includegraphics[width=1\textwidth]{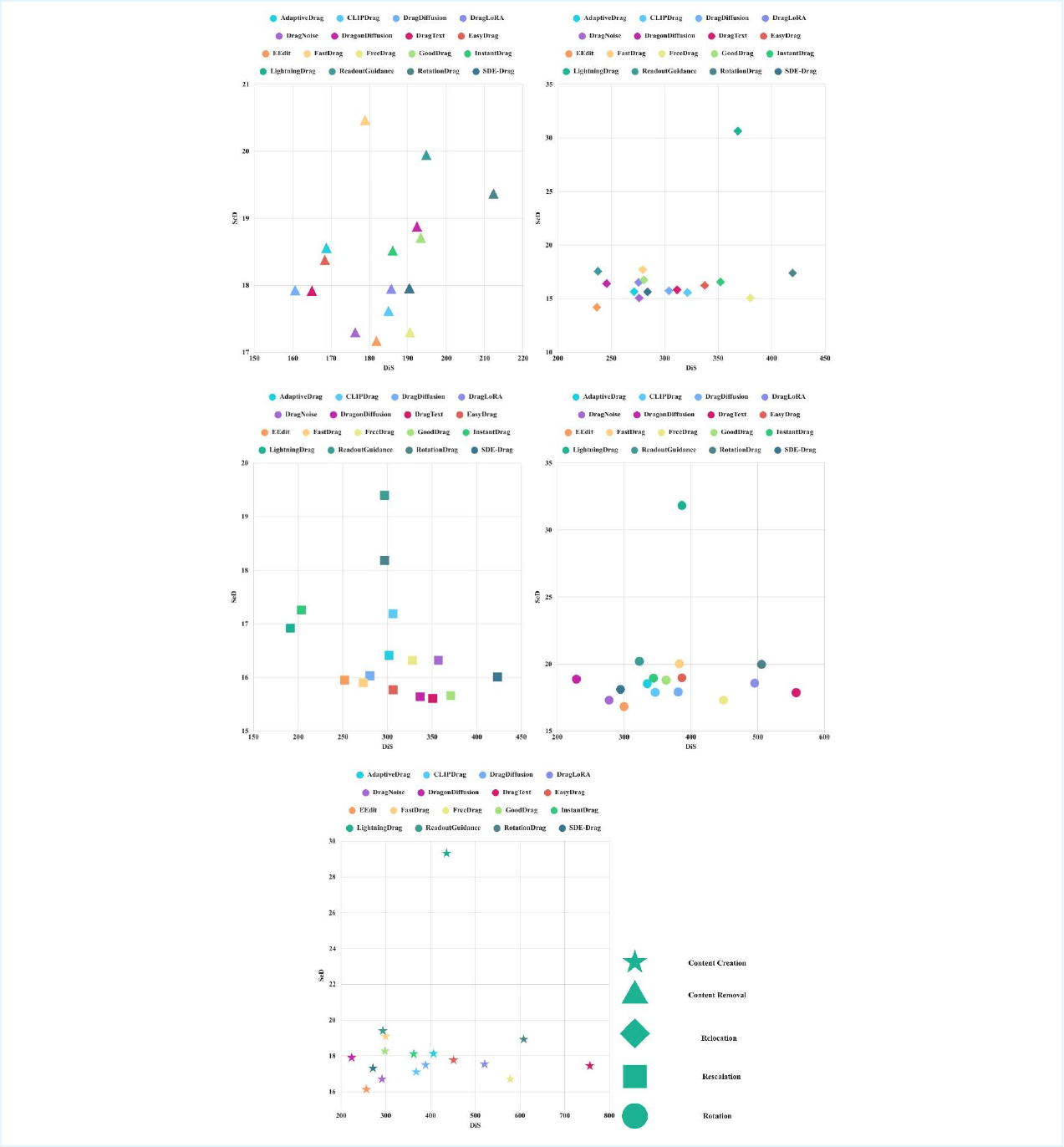}
  \caption{Per Task Metrics}
  \label{fig: perTaskMetrics}
\end{figure*}

In this subsection, we evaluate all models across the five drag task types using the most relevant generated-image–ground-truth similarity metrics. The corresponding results are presented in Figure \ref{fig: perTaskMetrics}. Unlike the aggregated results reported in Table \ref{tab:our_metrics}, the metrics here are computed without removing outliers and are scaled by a factor of 100 for ease of interpretation. All metrics are descending, meaning that better-performing models lie closer to the origin in the score space.

As shown in Figure \ref{fig: qualitative} (top), DragonDiffusion \cite{mou2023dragondiffusion} performs strongly on the Content Creation task. A similar trend can be observed in Figure \ref{fig: qualitative_2} (middle), where the model also produces visually convincing results on the Relocation task.

For the Rescaliation task, the two models achieving the best scores in Figure \ref{fig: perTaskMetrics} —DragText \cite{choi2025dragtext} and LightningDrag \cite{shi2024lightningdrag}—also deliver relatively strong qualitative results, as can be seen in Figure \ref{fig: qualitative_2} (top) and Figure \ref{fig: qualitative} (bottom).

DragNoise \cite{liu2024drag} achieves notable performance on the Rotation task, consistent with the qualitative example in Figure \ref{fig: qualitative_2} (bottom). On the Content Removal task, although DragNoise \cite{liu2024drag} and DragText \cite{choi2025dragtext} are unable to completely close the deer’s mouth (Figure \ref{fig: qualitative}, middle), they preserve the surrounding visual structures more faithfully than the two models that do close the mouth—FastDrag \cite{zhao2024fastdrag} and SDE-Drag \cite{nie2023blessing}—resulting in more natural images.

\subsection{IPPS Metric: Ablation Analysis}
\begin{figure}
  \includegraphics[width=\linewidth]{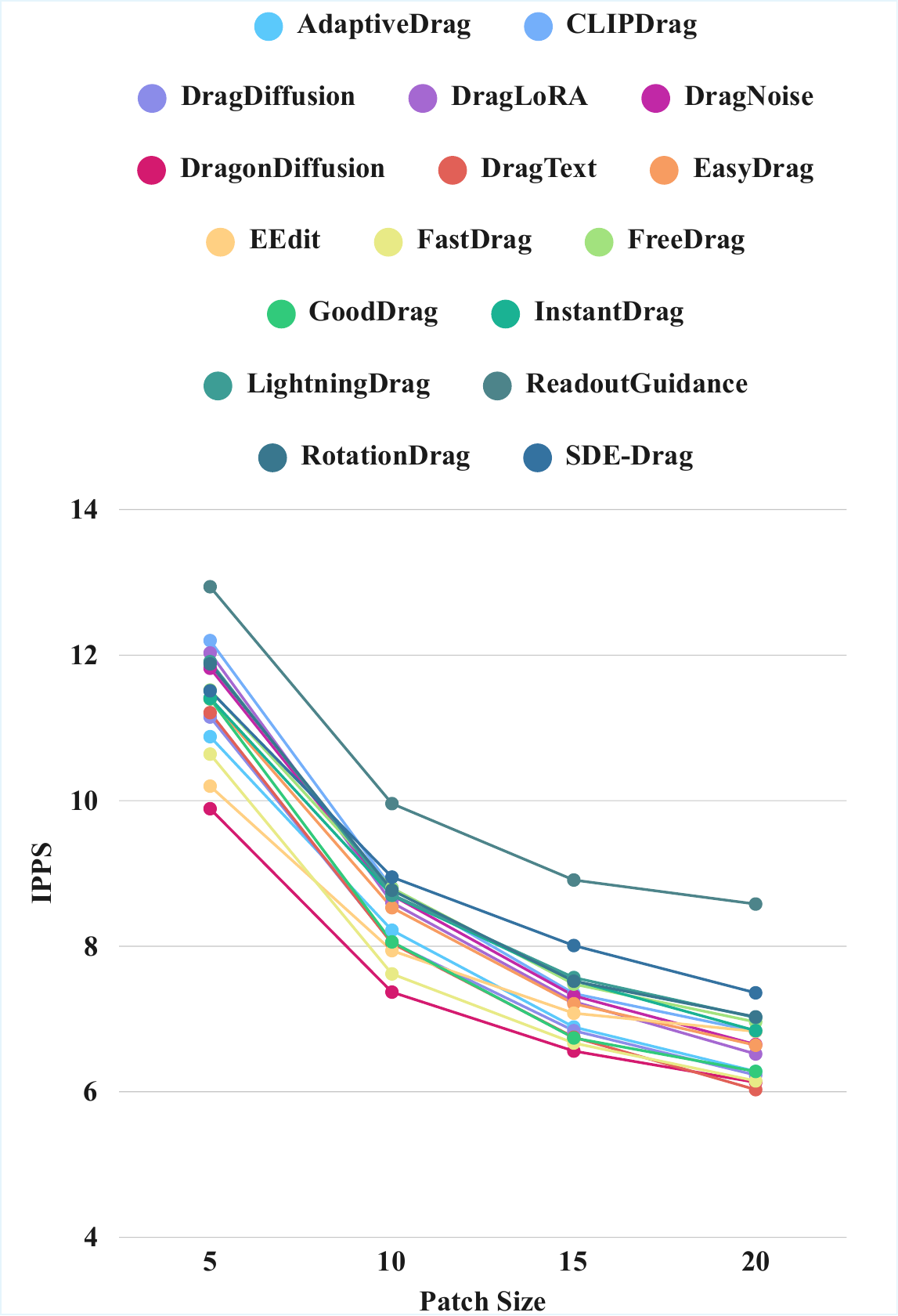}
  \caption{Ablation Study on IPPS Patch Size}
  \label{fig: ablationStudy}
\end{figure}

In this subsection, we perform an ablation study on the patch size used in the IPPS metric. In Table \ref{tab:our_metrics}, the patch radius was fixed at 5 pixels (corresponding to a 10 × 10 square patch). Here, we increase this radius to 10, 15, and 20 pixels, compute the IPPS metric using the same procedure as in Table \ref{tab:our_metrics}, and report the results in Figure \ref{fig: ablationStudy}. As shown, enlarging the patch size leads to higher similarity between the generated image and its ground truth within each patch. However, the ranking of the models remains largely unchanged: models that perform well under smaller patch sizes continue to perform well under larger ones.

\end{document}